\documentclass[journal,twoside,web]{ieeecolor}
\let\labelindent\relax
\usepackage{tmi}
\usepackage{amsmath,amssymb,amsfonts}
\usepackage{color}
\usepackage{hyperref}
\usepackage{cite}
\usepackage{algorithm}
\usepackage{url}
\usepackage{algorithmic}
\usepackage{graphicx}
\usepackage{textcomp}
\usepackage{enumitem}
\usepackage{bm}
\usepackage{makecell}
\usepackage{multirow}
\usepackage{bbding}
\usepackage{subfig}
\usepackage{booktabs}
\usepackage{threeparttable}
\def\BibTeX{{\rm B\kern-.05em{\sc i\kern-.025em b}\kern-.08em
    T\kern-.1667em\lower.7ex\hbox{E}\kern-.125emX}}
\begin{document}
%\title{Structurally-Guided Generalized Few-Shot Instance Segmentation for Annotation-Efficient Nucleus Instance Segmentation in Histopathology Images}
\title{Few-Shot Learning for Annotation-Efficient Nucleus Instance Segmentation}
\author{Yu Ming, Zihao Wu, Jie Yang, Danyi Li, Yuan Gao, Changxin Gao, Gui-Song Xia, Yuanqing Li \IEEEmembership{Fellow, IEEE}, Li Liang and Jin-Gang Yu 
\thanks{Yu Ming, Zihao Wu, Yuanqing Li and Jin-Gang Yu are with School of Automation Science and Engineering,~South China University of Technology,~Guangzhou 510641, China; Yuanqing Li and Jin-Gang Yu are also with Pazhou Laboratory, Guangzhou 510335, China. (E-mail: aumy@mail.scut.edu.cn;~auwzh@mail.scut.edu.cn;~auyqli @scut.edu.cn;~jingangyu@scut.edu.cn).}
\thanks{Jie Yang is with Department of Pathology, Zhujiang Hospital, Southern Medical University, Guangzhou 510280, China. (Email: tongt315@163.com).}
\thanks{Danyi Li and Li Liang are with Department of Pathology, Nanfang Hospital, Southern Medical University, Guangzhou 510515, China. (E-mail: lidanyi26@163.com; lli@smu.edu.cn).}
\thanks{Yuan Gao is with School of Electronic Information, Wuhan University, Wuhan 430072, China. (E-mail: yuangaoeis@whu.edu.cn).}
\thanks{Changxin Gao is with School of Artificial Intelligence and Automation, Huazhong University of Science and Technology, Wuhan 430074, China. (E-mail: cgao@hust.edu.cn).}
\thanks{Gui-Song Xia is with School of Computer Science, Wuhan University, Wuhan 430072, China. (E-mail: guisong.xia@whu.edu.cn).}
\thanks{Yu Ming and Zihao Wu contributed equally to this work. Corresponding author: Jin-Gang Yu and Li Liang.}}

\maketitle

\begin{abstract}
Nucleus instance segmentation from histopathology images suffers from the extremely laborious and expert-dependent annotation of nucleus instances. As a promising solution to this task, annotation-efficient deep learning paradigms have recently attracted much research interest, such as weakly-/semi-supervised learning, generative adversarial learning, etc. In this paper, we propose to formulate annotation-efficient nucleus instance segmentation from the perspective of few-shot learning (FSL). Our work was motivated by that, with the prosperity of computational pathology, an increasing number of fully-annotated datasets are publicly accessible, and we hope to leverage these external datasets to assist nucleus instance segmentation on the target dataset which only has very limited annotation. To achieve this goal, we adopt the meta-learning based FSL paradigm, which however has to be tailored in two substantial aspects before adapting to our task. First, since the novel classes may be inconsistent with those of the external dataset, we extend the basic definition of few-shot instance segmentation (FSIS) to generalized few-shot instance segmentation (GFSIS). Second, to cope with the intrinsic challenges of nucleus segmentation, including touching between adjacent cells, cellular heterogeneity, etc., we further introduce a structural guidance mechanism into the GFSIS network, finally leading to a unified Structurally-Guided Generalized Few-Shot Instance Segmentation (SGFSIS) framework. Extensive experiments on a couple of publicly accessible datasets demonstrate that, SGFSIS can outperform other annotation-efficient learning baselines, including semi-supervised learning, simple transfer learning, etc., with comparable performance to fully supervised learning with less than 5$\%$ annotations.
\end{abstract}


\section{Introduction}
\label{sec:intro}

\IEEEPARstart{N}ucleus instance segmentation, which aims to segment and classify individual cell nuclei from histopathology images, serves as a preliminary step towards many computational pathology tasks~\cite{abels2019computational}, such as quantitative characterization of tumor micro-environment~\cite{yuan2012quantitative}, immunohistochemical scoring~\cite{kapil2021domain,qaiser2019learning}, prognosis prediction\cite{gurcan2009histopathological}, etc. One primary challenge with nucleus instance segmentation is that, the annotation is extremely laborious and expertise-dependent, and hence only limited annotations can be acquired for training. 

To tackle this challenge, several annotation-efficient learning paradigms have been investigated in the literature. For example, {semi-supervised learning} takes advantage of abundant unlabeled data, in addition to limited labeled data, to boost the performance~\cite{li2020self, wu2022cross, jin2022semi}.  Generative adversarial learning~\cite{hou2019robust,mahmood2019deep} exploits Generative Adversarial Network (GAN) to synthesize labeled samples in order to augment the labeled training set. With the prosperity of computational pathology in recent years, an increasing number of fully-annotated datasets are publicly accessible~\cite{graham2019hover,gamper2019pannuke,verma2021monusac2020}. One natural and promising idea is to leverage these already existing datasets to facilitate the model learning on the target dataset. A representative annotation-efficient learning paradigm of this sort is domain adaptation (DA)~\cite{liu2020unsupervised, li2022domain, yang2021minimizing}. Nevertheless, an inherent limitation with DA is that, it assumes by definition that the classes of the target dataset are exactly identical to those of the external dataset, which is impractical in many application scenarios.

In this paper, we introduce a \textit{Structurally-Guided Generalized Few-Shot Instance Segmentation (SGFSIS)} framework for nucleus instance segmentation in hematoxylin-and-eosin (H\&E) stained histopathology images with limited annotations. First, we propose to formulate the task of nucleus instance segmentation, given a target dataset with limited annotations and an external dataset with full annotations, from the perspective of few-shot learning (FSL)~\cite{wang2020generalizing,vinyals2016matching,lang2023base}. Following the meta-learning paradigm, we first meta-train a few-shot instance segmentation (FSIS) model by episode sampling over the external dataset, and then fine-tune the model by using the limited annotations over the target dataset. Second, unlike DA~\cite{liu2020unsupervised, li2022domain, yang2021minimizing} which assumes the target classes to be exactly identical to those of the external dataset, or FSL which on the other hand assumes the two class sets to be strictly disjoint, we extend the basic definition of FSIS to generalized few-shot instance segmentation (GFSIS) which flexibly allows for the two class sets to partially overlap with each other. Third, in order to better conquer the inherent challenges with nucleus segmentation, like  touching instances and cellular heterogeneity, we particularly design a structural guidance mechanism in the GFSL network, i.e., exploiting the support set to modulate the structure prediction of the query, including foreground mask, boundary and centroid. To validate the proposed SGFSIS framework, we carry out extensive experiments on several public datasets, including ConSep~\cite{graham2019hover}, PanNuke~\cite{gamper2019pannuke}, MoNuSAC~\cite{verma2021monusac2020} and Lizard ~\cite{graham2021lizard}. The experimental results reveal that, the proposed method can outperform other annotation-efficient learning paradigms, including semi-supervised learning, simple transfer learning, etc., which achieves comparable performance to fully supervised learning with less than 5$\%$ annotations.

To sum up, the major contributions of our work are as follows:
\begin{itemize}
    \item We formulate annotation-efficient nucleus instance segmentation from the perspective of FSL, which has not been explored yet in the literature to our knowledge.
    \item We develop the SGFSIS framework to implement FSL-based annotation-efficient nucleus instance segmentation.
    \item We introduce an effective structural guidance mechanism into SGFSIS to improve the segmentation of adjacent touching nuclei.
\end{itemize}

\section{Related Work}
\label{sec:relatedwork}

%In this section, the closely related work in the literature is briefly reviewed.%, including deep learning based nucleus segmentation, annotation-efficient nucleus segmentation, few-shot instance segmentation in computer vision and few-shot learning for medical image segmentation.

\subsection{Fully-Supervised Nucleus Segmentation}

Fully-supervised nucleus segmentation methods can be categorized into traditional methods and deep learning based methods, and here we only focus on the closely relevant latter sub-category.

Some authors focus on how to effectively tailor U-Net \cite{ronneberger2015u} for the task of nucleus segmentation. Raza et al. \cite{raza2019micro} proposed Micro-Net to segment cells, nuclei or glands, which trains at multiple resolutions of the input image and connects the intermediate layers for better localization and context. Qu et al. \cite{qu2019improving} proposed FullNet, which maintains full resolution feature maps in U-Net to improve the localization accuracy.

One key challenge is to accurately segment touching nuclei, for which the majority of works seeks to predict by deep learning certain clues so as to effectively represent nucleus instances. In the post-processing step, the predicted clues can then be taken to extract guidance markers and initiate a marker-guided watershed procedure to fulfill accurate nucleus segmentation. Along this line, Xing et al.~\cite{xing2015automatic} combined a nucleus mask derived from a CNN-based patch classifier and a shape prior model (represented as nucleus boundary) for accurate nucleus segmentation. Chen et al. ~\cite{chen2017dcan} presented DCAN to utilize nucleus boundary as an additional clue. Ke et al.~\cite{ke2023clusterseg} developed the ClusterSeg framework featured by a branch to predict the clustered boundaries between touching nuclei. Other representative works include CIA-Net~\cite{zhou2019cia}, Triple U-Net~\cite{zhao2020triple}, etc. 

%Zhou et al. \cite{zhou2019cia} introduced CIA-Net which deploys a multi-level
%information aggregation module with bi-directional connections between the nucleus mask decoder and the spatial contour decoder.  Zhao et al.~\cite{zhao2020triple} proposed the Triple U-Net consisting of an RGB channel to predict nucleus mask, an Hematoxylin channel to predict boundary and a segmentation branch to integrate these two.

Beside nucleus boundary, the distance map is also widely utilized as an additional clue to enhance nucleus segmentation ~\cite{naylor2018segmentation} in the literature. Naylor et al. ~\cite{naylor2018segmentation} formulated touching nucleus segmentation as a regression task of the distance map. As a very impactful work on nucleus instance segmentation, Granhan et al. ~\cite{graham2019hover} presented Hover-Net which simultaneously predicts the vertical and horizontal distances from each nucleus pixel to its centroid. Schmidt et al. [18] proposed the StarDist method which predicts the centroid probability map as well as the distance map along a couple of pre-defined directions. There are also some other successful approaches, including CDNet ~\cite{he2021cdnet}, CPP-Net ~\cite{chen2023cpp}, TopoSeg ~\cite{he2023toposeg}, etc.

While these works address fully-supervised nucleus segmentation, our work is concerned with the scenario where only limited annotation is available.  

\subsection{Annotation-Efficient Nucleus Segmentation}
Several annotation-efficient learning paradigms, including data augmentation \cite{hou2019robust, mahmood2019deep}, semi-supervised learning \cite{li2020self, wu2022cross, jin2022semi, ke2023clusterseg}, domain adaptation \cite{liu2020unsupervised, li2022domain, yang2021minimizing}, weakly-supervised learning \cite{qu2020weakly, lin2024bonus, zhou2023cyclic}, etc., have been investigated in the literature in order to conquer the scarcity of annotation in nucleus segmentation.

Data augmentation based methods deploy GAN~\cite{hou2019robust} and its variants~\cite{mahmood2019deep} to synthesize nucleus instances so as to alleviate the shortage of labeled training samples. Semi-supervised learning based methods~\cite{li2020self, wu2022cross, jin2022semi, ke2023clusterseg} leverage abundant unlabeled data, in addition to limited amount of labeled data, to boost the performance of nucleus instance segmentation, most typically following the Teacher-Student framework~\cite{jin2022semi}. Similar to our work, domain adaptation based methods~\cite{liu2020unsupervised, li2022domain, yang2021minimizing} also take advantages of external labeled datasets to assistant nucleus instance on the target dataset. Weakly-supervised learning based methods consider the much less expensive annotations, like point annotation~\cite{qu2020weakly,lin2024bonus}, image-level label~\cite{zhou2023cyclic}, etc.  

Recently, Lou et al.~\cite{lou2022pixel} suggested an interesting framework which integrates data augmentation, semi-supervised learning and a selection strategy to determine which patches are most value to be annotated. Han et al. \cite{han2022meta} proposed a meta multi-task learning model to reduce the data dependency of nucleus instance segmentation.

Our work investigates a novel nucleus segmentation paradigm based on FSL.

\begin{figure*}[!t]
	\centering
	\includegraphics[width = \linewidth]{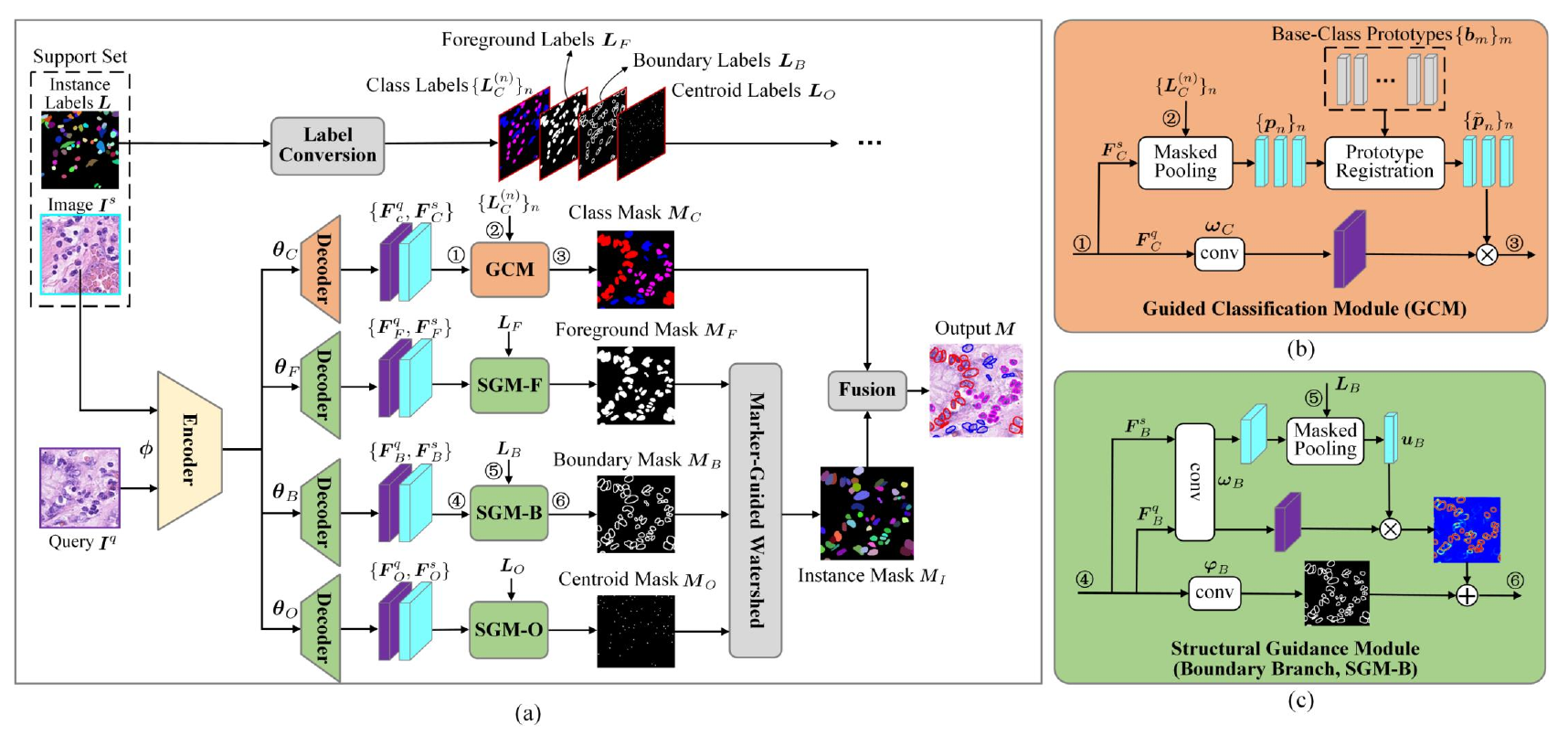}
	\caption{Overview of the Structurally-Guided Generalized Few-Shot Instance Segmentation (SGFSIS) framework, with (a) the overall network architecture and the details of (b) the Guided Classification Module (GCM) and (c) the Structural Guidance Module (SGM), where we take SGM-B as an example for presentation while SGM-F/O have an exactly identical structure.}
\label{fig:framework}
\end{figure*}

\subsection{Few-Shot Instance Segmentation in Computer Vision}
Few-shot instance segmentation (FSIS) aims to learn an instance segmentation model with a large labeled training dataset of base classes and apply it to the test set of novel classes given only a few labeled novel-classes samples ~\cite{fan2020fgn, wang2022dynamic,kohler2023few}. Some works further extend the basic FSIS setting to generalized FSIS in order to allievate the forgetting of base classes~\cite{fan2021generalized,ma2023digeo}. It is an active research topic in computer vision in recent years, and for a more comprehensive review, we refer the readers to ~\cite{kohler2023few}. These general FSIS approaches cannot be directly deployed to the task of nucleus segmentation.

\subsection{Few-Shot Learning for Medical Image Segmentation}
FSL techniques have been exploited to the segmentation of other modalities of medical images~\cite{roy2020squeeze, cui2020unified, tang2021recurrent, feng2023learning}, such as CT, MRI, X-ray, etc. Following the standard small-sample learning setting, Roy et al.~\cite{roy2020squeeze}  proposed a ``channel squeeze \& spatial excitation" module to enhance the interaction between the support branch and the query branch for CT image segmentation. Cui et al.~\cite{cui2020unified} proposed a unified framework for generalized low-shot medical image segmentation based on distance metric learning with application to MRI and CT image segmentation. FSL so far has rarely been explored in the task of nucleus instance segmentation.

\section{Method}
\label{sec:approach}
%In this section, we start with notations and problem statement, followed by describing the proposed SGFSIS framework and its key components.
\subsection{Problem Statement}

Given a training set of  histopathology images $\mathcal{D}$ (the \textit{target dataset}), our task is to learn from it a nucleus
instance segmentation model that can segment and classify every nucleus individual belonging to the classes of interest $\mathcal{C}$ (also called the \textit{novel classes} as described later). As precise instance
labels are extremely expensive, it is common in practice that only a very small subset $\mathcal{D}_l \subset \mathcal{D}$ is labeled while leaving the rest majority $\mathcal{D} \setminus \mathcal{D}_l$ unlabeled. 

Our major contribution is to introduce few-shot learning (FSL) into the task of nucleus
instance segmentation as a novel annotation-efficient learning paradigm. Note that a prerequisite is the availability of an \textit{external dataset} $\mathcal{D}^{\text{base}}$, which is completely labeled with nucleus instances belonging to the \textit{base classes} $\mathcal{C}^{\text{base}}$. Fortunately, this is feasible since an increasing number of public datasets emerge with the rapid advances on computational pathology in recent years. Our goal is to leverage the external dataset $\mathcal{D}^{\text{base}}$ of the base classes $\mathcal{C}^{\text{base}}$ to boost the model training over the target dataset $\mathcal{D}$ of the classes $\mathcal{C}$, hopefully under the FSL framework. 

Following the terminology of FSL, we regard the small labeled training subset $\mathcal{D}_l$ as the \textit{support set} $\mathcal{S}$, i.e.,  $\mathcal{S} \triangleq \mathcal{D}_l$, and its classes $\mathcal{C}$ as the \textit{novel classes}. Let us further suppose $|\mathcal{C}| = N$, $|\mathcal{C}^{\text{base}}| = M$ and there exist $|\mathcal{S}| = K$ labeled images, {denoted by} $\mathcal{S} = \{ (\boldsymbol{I}_k^s, \boldsymbol{L}_k)\}_{k=1}^K$ where $\boldsymbol{I}_k^s$ is an image and $\boldsymbol{L}_k$ the corresponding class mask. By using both $\mathcal{D}^{\text{base}}$ and $\mathcal{S}$, our task is to learn a conditional model $f(\boldsymbol{I}^q|\mathcal{S})$ so that, for a testing image $\boldsymbol{I}^q$ (or called a \textit{query}), it can segment and classify every nucleus individual belonging to the novel classes $\mathcal{C}$. Conventionally, such a setting is referred to as an \textit{$N$-way $K$-shot few-shot instance segmentation (FSIS)} task.
% The classes $\mathcal{C}^{\text{base}}$ of the external dataset $\mathcal{D}^{\text{base}}$ are regarded as the base classes. 

Particularly, the basic definition of FSL above requires the base classes and the novel classes to be strictly disjoint, i.e., $\mathcal{C}^{\text{base}} \cap \mathcal{C} = \phi$. Nevertheless, we cannot guarantee the available external dataset always has totally different (non-overlapping) classes from the target dataset in  practice. To better adapt to realistic applications, we extend the problem setting of FSIS above by allowing the two class sets to overlap with each other, i.e.,  $\mathcal{C}^{\text{base}} \cap \mathcal{C} \neq \phi$, leading to \textit{generalized few-shot instance segmentation (GFSIS)}.

\subsection{The Proposed SGFSIS Framework} 
\label{sec:SGFSIS-framework}
We propose a Structurally-Guided  Generalized Few-shot Instance Segmentation (SGFSIS) framework to implement the FSIS task stated above, as illustrated in Fig.~\ref{fig:framework}. There are three major considerations in developing SGFSIS as below.

\vspace{6pt}
\subsubsection{Basic Network for Nucleus Instance Segmentation} 
Inspired by the success of previous works on fully-supervised nucleus instance segmentation, like Hover-Net~\cite{graham2019hover}, DCAN~\cite{chen2017dcan}, etc., our basic network consists of four branches, namely the classification branch (CB), the foreground branch (FB), the boundary branch (BB) and the centroid branch (OB), as shown in Fig.~\ref{fig:framework}(a). CB predicts a classification mask which assigns a class label to each pixel (which essentially performs semantic segmentation). FB, BB and OB predict structural information about nuclei, each yielding a mask that quantifies the probability of each pixel being nucleus foreground, boundary and centroid respectively. The three masks obtained by FB, BB and OB are taken together to initiate a \textit{marker-guided watershed} algorithm (as detailed in Section~\ref{subsec:watershed}), which then generates a class-agnostic instance mask. This instance mask is finally fused with the classification mask to yield the output instance segmentation mask. The four branches adopt a typical encoder-decoder structure, which share an encoder (parameterized by $\boldsymbol{\phi}$) while each having a separate decoder (parameterized by $\boldsymbol{\theta}_C$, $\boldsymbol{\theta}_F$, $\boldsymbol{\theta}_B$ and $\boldsymbol{\theta}_O$ respectively). In each branch, the support set and the associated labels are taken to guide the prediction of the corresponding mask.

\vspace{6pt}
\subsubsection{Guidance Mechanisms} 
Vital to our SGFSIS are the guidance mechanisms of each branch. For the CB branch, we design the \textit{Guided Classification Module (GCM)}, as shown in Fig.~\ref{fig:framework}(b) and detailed in Section~\ref{subsec:gcm}. And for the FB, BB and OB branches, we design the \textit{Structural Guidance Modules (SGMs)}, named SGM-F, SGM-B and SGM-O respectively, as shown in Fig.~\ref{fig:framework}(c) and detailed in Section~\ref{subsec:sgm}. 

\vspace{6pt}
\subsubsection{Training Strategy} 
Basically we follow a three-stage meta-learning scheme to train the model, which includes pre-training, episode sampling based meta-training and a fine-tuning procedure, as detailed in Section~\ref{subsec:trainingstr}.

\subsection{Guided Classification Module (GCM) } \label{subsec:gcm}

The proposed GCM module is shown in Fig.~\ref{fig:framework}(b). For clarity, let us suppose the support set includes only one labeled image, i.e., $K=1$ and $\mathcal{S} = \{(\boldsymbol{I}^s, \boldsymbol{L})\}$. We first perform label conversion to split the overall instance label image $\boldsymbol{L}$ into several channels, including the classification labels $\{\boldsymbol{L}_C^{(n)}\}_{n=1}^N$ with $\boldsymbol{L}_C^{(n)}$ being that of the  class $n$, the foreground label $\boldsymbol{L}_F$, the boundary label $\boldsymbol{L}_B$ and the centroid label $\boldsymbol{L}_O$.  Then both the support-set image $\boldsymbol{I}^s$ and the query image $\boldsymbol{I}^q$ are fed into the encoder $\boldsymbol{\phi}$ and the decoder $\boldsymbol{\theta}_C$ to get the feature maps  $\boldsymbol{F}_C^s$ and $\boldsymbol{F}_C^q$ respectively. The classification labels $\boldsymbol{L}_C^{(n)}$ belonging to the class $n$ is taken to perform the \textit{masked pooling} operation over the feature maps $\boldsymbol{F}_C^s$ to get the novel-class prototypes $\{\boldsymbol{p}_n\}_{n=1}^N$ as follows  
\begin{equation}
{\boldsymbol{p}}_n \leftarrow \text{GAP} \left( \boldsymbol{F}_C^s \odot \boldsymbol{L}_C^{(n)} \right), 
\label{eq:novelproto}
\end{equation}
where $\odot$ stands for the masking operator and $\text{GAP}(\cdot)$ the operator of global average pooling. Notice that if there exist $K > 1$ labeled images in the support set, we average the corresponding $\boldsymbol{p}_n$'s over the multiple images. 

In order to better transfer knowledge about the external dataset $\mathcal{D}^{\text{base}}$ to boost the target task, we learn from $\mathcal{D}^{\text{base}}$ a set of $M$ base-class prototypes $\{\boldsymbol{b}_m\}_{m=1}^M$ \cite{tian2022generalized, liu2023learning}. Concretely, starting from a set of randomly initialized vectors, every image in $\mathcal{D}^{\text{base}}$ is fed through the encoder $\boldsymbol{\phi}$ and the decoder $\boldsymbol{\theta}_C$ to get the feature maps $\boldsymbol{F}$ (Note that we share the encoder $\boldsymbol{\phi}$ and the decoder $\boldsymbol{\theta}_C$ in the CB branch for the base-class prototype learning), which are then converted into the classification masks $\{\boldsymbol{S}^{(m)}\}_{m=1}^M$ by
\begin{equation}
  \boldsymbol{S}^{(m)} \leftarrow \boldsymbol{F} \otimes {\boldsymbol{b}}_m,
  \label{eq:odot}
\end{equation}
\begin{equation}
  \boldsymbol{S}^{(m)} \leftarrow \text{softmax}_m \left\{ \boldsymbol{S}^{(m)} \right\},
\end{equation}
where the $\otimes$ operator in Eqn.~(\ref{eq:odot}) is concretely defined as
\begin{equation}
  \boldsymbol{S}^{(m)}(x,y) =  \cos \left[ \boldsymbol{F}(x,y), {\boldsymbol{b}}_m \right],
  \label{eq:baseproto}
\end{equation}
with $(x,y)$ being a two-dimensional pixel location and $\cos(\cdot)$ the cosine similarity measure (we follow such a definition throughout this paper). The predicted classification masks $\{\boldsymbol{S}^{(m)}\}_{m=1}^M$ is finally compared against the label image $\boldsymbol{L}$ to establish the training loss, for which we adopt the commonly-used pixel-wise cross-entropy loss.

Once we obtain the base-class prototypes $\{\boldsymbol{b}_m\}_{m=1}^M$, they are fused with the novel-class prototypes $\{\boldsymbol{p}_n\}_{n=1}^N$ to yield $\{\tilde{\boldsymbol{p}}_n\}_{n=1}^N$ by the following \textit{prototype registration} procedure
\begin{equation}
\tilde{\boldsymbol{p}}_{n} = 
\begin{cases}
\gamma_n \boldsymbol{p}_{n}+\left(1-\gamma_n\right)  \boldsymbol{b}_m, &  \text{if} \ {\exists}  m, \    \mathcal{C}^{\text{base}}_m = \mathcal{C}_n, \\
\boldsymbol{p}_{n}, & \text{otherwise}. 
\end{cases} 
\label{eq:protofuse}
\end{equation}
where $\gamma_n = \text{cos}(\boldsymbol{p}_{n}, \boldsymbol{b}_{m})$ measures the cosine similarity between
$\boldsymbol{p}_{n}$ and $\boldsymbol{b}_{m}$. Intuitively, Eqn.~(\ref{eq:protofuse}) suggests that, if the novel classes overlap with the base classes, the prototypes of the overlapped base classes will be used to update the corresponding novel-class prototypes. The obtained prototypes $\{\tilde{\boldsymbol{p}}_{n}\}_{n=1}^N$ are then used to generate the classification masks $\{ \boldsymbol{M}^{(n)}_C \}_{n=1}^N$ as follows
\begin{equation}
    \boldsymbol{M}^{(n)}_C \leftarrow \text{softmax}_n \left\{ \text{conv}_{\boldsymbol{\omega}_C}(
    \boldsymbol{F}_C^q) \otimes \tilde{\boldsymbol{p}}_n \right\},
\label{eq:clsguidance}
\end{equation}
where $\boldsymbol{\omega}_C$ is the learnable parameters of the convolution layer.

\begin{figure}[!t]
	\centering
	\includegraphics[width = 0.8\linewidth]{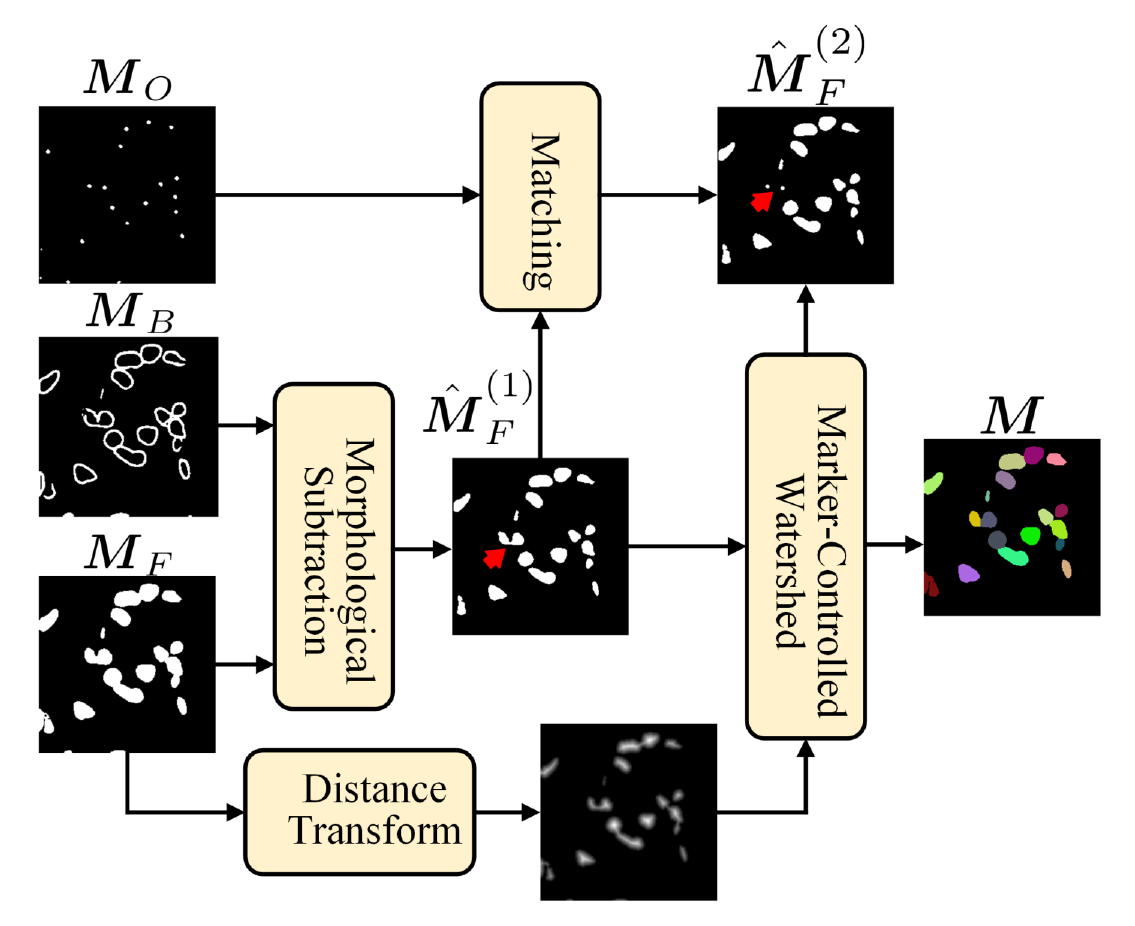}
	\caption{Pipeline of the Marker-Guided Watershed module.}
\label{fig:marker_guided_watershed}
\end{figure}

\subsection{Structural Guidance Modules (SGMs) }
\label{subsec:sgm}
The three structural guidance modules SGM-B, SGM-F and SGM-O share the same network structure and we just take SGM-B for an instance for description. As shown in Fig.~\ref{fig:framework}(c), the obtained feature maps $\boldsymbol{F}_B^s$ goes through a convolution layer $\boldsymbol{\omega}_B$, followed by taking the boundary label $\boldsymbol{L}_B$ to conduct a masked pooling operation and get a class-agnostic prototype $\boldsymbol{u}_B$ by
\begin{equation}
    \boldsymbol{u}_B \gets \text{GAP} \left( \text{conv}_{\boldsymbol{\omega}_B}(
    \boldsymbol{F}_B^s) \odot \boldsymbol{L}_B \right).  
\label{eq:proto}
\end{equation}
It is worth pointing out that, unlike for GCM in Eqn.~(\ref{eq:baseproto}), we do not integrate base-class information about the external dataset in calculating $\boldsymbol{u}_B$. This is because the prototype here is class-agnostic, and knowledge about the external dataset can be effectively transferred via the preceding meta-training procedure. The boundary mask $\boldsymbol{M}_B$ is then given by
\begin{equation}
\boldsymbol{M}_B \gets \text{conv}_{\boldsymbol{\omega}_B}(
    \boldsymbol{F}_B^q) \otimes \boldsymbol{u}_B   + \text{conv}_{\boldsymbol{\varphi_B}}(
    \boldsymbol{F}_B^s),
\label{eq:sgmMb}
\end{equation}
where $\boldsymbol{\omega}_B$ and $\boldsymbol{\varphi}_B$ are the parameters of the two convolution layers. Similarly, the parameters for the SGM-F and SGM-O modules are denoted by $\boldsymbol{\omega}_F$, $\boldsymbol{\varphi}_F$, $\boldsymbol{\omega}_O$ and $\boldsymbol{\varphi}_O$ respectively.

\subsection{Marker-Guided Watershed}
\label{subsec:watershed}
Once the foreground mask $\boldsymbol{M}_F$, the boundary mask $\boldsymbol{M}_B$ and the centroid mask $\boldsymbol{M}_O$ have been predicted by the SGM-F, SGM-B and SGM-O modules respectively, they are integrated to derive a marker map, which is then used to initiate a marker-controlled watershed procedure to generate the instance mask $\boldsymbol{M}_I$, as illustrated in Fig.~\ref{fig:marker_guided_watershed}.
First, we subtract $\boldsymbol{M}_B$ from $\boldsymbol{M}_F$ via the morphological
erosion operation, yielding an eroded foreground mask $\hat{\boldsymbol{M}}^{(1)}_F$. Second, we perform connected component labeling over $\hat{\boldsymbol{M}}^{(1)}_F$ and  ${\boldsymbol{M}}_O$ to get the sets of connected components  $\hat{\mathcal{A}}_F$ and $\mathcal{A}_O$. Third, we further refine $\hat{\boldsymbol{M}}^{(1)}_F$ by spatially matching between the connected components in $\hat{\mathcal{A}}_F$ and $\mathcal{A}_O$. For every connected component $\mathbf{g} \in \hat{\mathcal{A}}_F$, if it contains more than one connected component in $ {\mathcal{A}}_O$, we then use these multiple  connected components to replace  $\mathbf{g}$ (see those highlighted in red in Fig.~\ref{fig:marker_guided_watershed} as an example), or otherwise, we keep $\mathbf{g}$ unchanged. The refined foreground mask $\hat{\boldsymbol{M}}^{(2)}_F$ is then taken as the marker to guide a watershed procedure so as to obtain the instance mask  $\mathbf{M}_I$.

\subsection{Training Strategy} \label{subsec:trainingstr}

We basically follow the episode sampling based meta-learning paradigm~\cite{hospedales2021meta, tian2022generalized} to train the SGFSIS framework, whose parameters to be learnt  are summarized by $
\boldsymbol{\Omega} = \left\{ \boldsymbol{\phi}, \boldsymbol{\theta}_C, \boldsymbol{\theta}_F, \boldsymbol{\theta}_B, \boldsymbol{\theta}_O; 
\{\boldsymbol{b}_m\}_{m=1}^M;
\boldsymbol{\omega}_C; \boldsymbol{\omega}_F,
\boldsymbol{\varphi}_F; \boldsymbol{\omega}_B, 
\boldsymbol{\varphi}_B; \boldsymbol{\omega}_O,
\boldsymbol{\varphi}_O \right\}$.
Throughout this paper, we use the standard cross-entropy loss for classification and the DICE loss for dense prediction. For the encoder $\boldsymbol{\phi}$, we use the ResNet-50 backbone network ~\cite{he2016deep}. We adopt a three-stage training strategy as follows:

\subsubsection{Pre-training on $\mathcal{D}^{\text{base}}$} The first stage is to pre-train over $\mathcal{D}^{\text{base}}$ a part of the parameters in a fully-supervised manner. More precisely, we remove all the components related with the guidance mechanisms in the SGFSIS framework as illustrated in Fig.~\ref{fig:framework} while keeping the remainder, including: 1) the encoder  $\boldsymbol{\phi}$; 2) the decoder $\boldsymbol{\theta}_C$, the base-class prototypes $\{\boldsymbol{b}_m\}_{m=1}^M$ and the convolution layer $\boldsymbol{\omega}_C$ in the CB branch, which predicts the classification masks according the strategy of base-class prototype learning as detailed in Section~\ref{subsec:gcm}; 3) the decoders $\boldsymbol{\theta}_F$, $\boldsymbol{\theta}_B$ and $\boldsymbol{\theta}_O$, and the corresponding convolution layers $\boldsymbol{\omega}_F$, $\boldsymbol{\omega}_B$ and $\boldsymbol{\omega}_O$ in the three SGM modules respectively, which predict the corresponding masks. For every training image, its ground-truth instance labels after label conversion  are taken to supervise the corresponding mask prediction. The ResNet-50 encoder $\boldsymbol{\phi}$ is initialized by the one pre-trained on ImageNet, and the decoders are randomly initialized.

\subsubsection{Meta-training on $\mathcal{D}^{\text{base}}$} In the second stage, we construct the so-called episodes on the external dataset $\mathcal{D}^{\text{base}}$ to simulate the $N$-way $K$-shot GFSIS task on the target dataset, which are then taken to meta-train the entire model parameters $
\boldsymbol{\Omega}$. By our problem definition, a GFSIS task (episode) can be regarded as $\mathcal{T} = \{\textit{support set}, \textit{novel classes}, \textit{base classes}, \textit{query}\}$. 
Mostly following~\cite{tian2022generalized}, in order to construct a GFSL episode $\mathcal{T}$, we randomly sample from the external dataset a number of images (which is actually a mini-batch and the number equals to the batch size). Then we randomly take a half from these images as the support set and the corresponding classes as the novel classes, and the other half of the images as the queries. Further, from the novel classes we randomly select a half as the base classes. For these base classes, we take the corresponding entries from the base-class prototypes obtained in the first stage to be the base-class prototypes of this episode. Episodes sampled in this way are then used to meta-train the entire parameters $\boldsymbol{\Omega}$.

\subsubsection{Fine-tuning on $\mathcal{S}$} In the third phase, we first use the labeled images in the support set $\mathcal{S}$ to upgrade the prototypes $\{\boldsymbol{p}_n\}_{n=1}^N$ according to Eqn.~(\ref{eq:novelproto}) and the class-agnostic prototypes 
$\{\boldsymbol{u}_B, \boldsymbol{u}_F, \boldsymbol{u}_O\}$ according to Eqn.~(\ref{eq:proto}). Then, we alternately take every image in $\mathcal{S}$ as the query and its corresponding labels as the ground truth to fine-tune the parameters $
\boldsymbol{\Omega}$ in a fully-supervised way.  

\begin{table*}[]
\centering
\begin{footnotesize}
\caption{Basic information about the four datasets. Notice that the TCGA* dataset is collected from tens of different centers, and UHCW is short for University Hospitals Coventry and Warwickshire. }
\renewcommand{\arraystretch}{1.2}
\begin{tabular}{cccccc}
\Xhline{1pt}
\textbf{Dataset} & \textbf{Tissues} & \textbf{Centers} & \textbf{Magnification} & \textbf{Class Labels} & \textbf{\#Images (Training/Testing)} \\ 
\Xhline{1pt}
CoNSeP  & Colon & UHCW & 40$\times$ & INF, EPI, SPS, MIS & 1748/224  \\ 
\hline
MoNuSAC & \begin{tabular}[c]{@{}l@{}}Prostate, Breast,\\ Kidney, Lung\end{tabular} & TCGA* & 40$\times$  & EPI, LYM, NEU, MAC& 2031/858 \\ 
\hline
PanNuke & 19 Organs & TCGA*, UHCW & 40$\times$ &  NEO, INF,CON, DEA, EPI & 5179/2722 \\ 
\hline
Lizard  & Coloretal & \begin{tabular}[c]{@{}l@{}} UHCW, TCGA*, \\ 4 Chinese hospitals\end{tabular} & 20$\times$ & EPI, LYM, PLA, NEU, EOS, CON & 3997/984 \\ 
\Xhline{1pt}
\end{tabular}
\label{tab:datasetinfo}
\end{footnotesize}
\end{table*}

\begin{figure*}[!t]
	\centering
	\includegraphics[width = 0.85\linewidth]{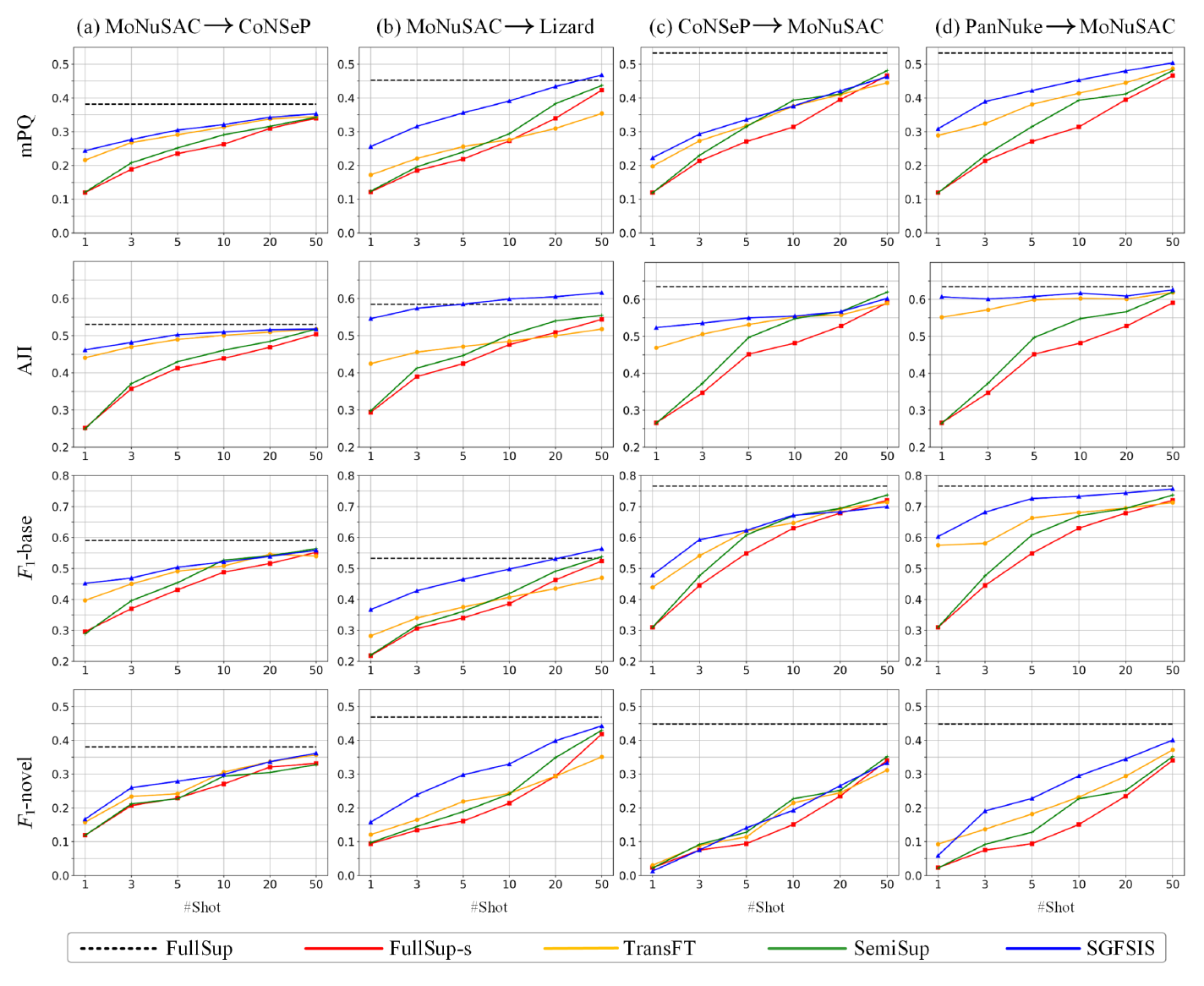}
	\caption{Plots of the quantitative results obtained by our SGFSIS and the three baselines in terms of mPQ, AJI, $F_1$-base and $F_1$-novel, with the shot number varying from 1 to 50 over the four different dataset settings.}
\label{fig:quanti-results}
\end{figure*}

\section{Experimental Setup}
\label{sec:experiment}
%In this section, we describe our experimental settings, including datasets, performance metrics, implementation details and methods for comparison.

\subsection{Datasets}
\vspace{-2pt}
 We use four publicly available datasets for our experiments, termed as \textbf{ConSep}~\cite{graham2019hover}, \textbf{PanNuke}~\cite{gamper2019pannuke,gamper2020pannuke}, \textbf{MoNuSAC}~\cite{verma2021monusac2020} and \textbf{Lizard} ~\cite{graham2021lizard} respectively. These datasets all contain H\&E-stained histopathology images of nuclei with accurate pixel-level instance labels, which cover 12 cell types (class labels), including inflammatory (INF), epithelial (EPI), spindle-shaped (SPS), lymphocyte (LYM), neutrophil (NEU), macrophage (MAC), neoplastic (NEO), connective (CON), dead (DEA), plasma (PLA), eosinophil (EOS) and miscellaneous (MIS). Basic information about these four datasets, including tissue types, magnification factors, class labels, image numbers, etc., are summarized in Table~\ref{tab:datasetinfo}. 
 
 We use a uniform image size of 256 $\times$ 256 pixels for our experiments. \textbf{ConSep} originally consists of 41 images sized by 1000 $\times$ 1000 and \textbf{MoNuSAC} consists of 310 images of varying sizes, so we follow the protocol of~\cite{graham2019hover} to crop 256 $\times$ 256 images from these larger ones, leading to 1748 and 2031 images respectively.  

 We consider the following settings of the external dataset $\mathcal{D}^{\text{base}}$ and the target dataset $\mathcal{D}$, referred to as 
 \textbf{MoNuSAC $\rightarrow$ ConSep}, \textbf{MoNuSAC $\rightarrow$ Lizard}, 
 \textbf{ConSep $\rightarrow$ MoNuSAC} and  \textbf{PanNuke $\rightarrow$ MoNuSAC}, where MoNuSAC $\rightarrow$ ConSep means taking MoNuSAC as the external dataset and ConSep as the target dataset, and so are the other three settings.

\begin{table*}[!htb]
\scriptsize
  \centering
  \begin{scriptsize}
  \caption{Quantitative results obtained by our SGFSIS and the three baselines in terms of mPQ, AJI, $F_1$-base and $F_1$-novel, with the shot number varying from 1 to 50 over the four different dataset settings. The best result under each setting is highlighted in \textbf{bold}.}
    \setlength{\tabcolsep}{1.2mm}{
    \begin{tabular}{clcccccccccccccccc}
    \Xhline{1pt}
    
    \multirow{2}{*}{\textbf{\#Shots}} & \multirow{2}{*}{\textbf{Methods}}  
    & \multicolumn{4}{c}{\textbf{MoNuSAC}$\rightarrow$\textbf{CoNSeP}} 
    & \multicolumn{4}{c}{\textbf{MoNuSAC}$\rightarrow$\textbf{Lizard}} 
    & \multicolumn{4}{c}{\textbf{CoNSeP}$\rightarrow$\textbf{MoNuSAC}} 
    & \multicolumn{4}{c}{\textbf{PanNuke}$\rightarrow$\textbf{MoNuSAC}} 
    \\

    \cmidrule(lr){3-6}
    \cmidrule(lr){7-10}
    \cmidrule(lr){11-14}
    \cmidrule(lr){15-18}
    
    {} & {} & mPQ & AJI &  $F_1$-base  & $F_1$-novel & mPQ & AJI &  $F_1$-base & $F_1$-novel & mPQ & AJI &  $F_1$-base  & $F_1$-novel & mPQ & AJI &  $F_1$-base  &$F_1$-novel\\
    
    % \cline{1-{18}}
    \Xhline{1pt}
     %newrow
     \Gape[3pt]
      %newrow
     
    \multirow{4}[0]{*}{1} 
          & FullSup-s      & 0.120 & 0.252 & 0.296 & 0.119
                           & 0.122 & 0.294 & 0.218 & 0.094
                           & 0.120 & 0.266 & 0.310 & 0.023
                           & 0.120 & 0.266 & 0.310 & 0.023
                           \\
          & TransFT        & 0.216 & 0.441 & 0.397 & 0.158
                           & 0.172 & 0.425 & 0.282 & 0.121
                           & 0.198 & 0.469 & 0.439 & \textbf{0.030}
                           & 0.289 & 0.552 & 0.575 & \textbf{0.093}
                           \\
          & SemiSup        & 0.121 & 0.249 & 0.289 & 0.119
                           & 0.124 & 0.298 & 0.220 & 0.097
                           & 0.119 & 0.265 & 0.310 & 0.022
                           & 0.119 & 0.265 & 0.310 & 0.022
                           \\
          & SGFSIS (ours)   & \textbf{0.244} & \textbf{0.462} & \textbf{0.452} & \textbf{0.167}
                           & \textbf{0.256} & \textbf{0.546} & \textbf{0.367} & \textbf{0.158}
                           & \textbf{0.223} & \textbf{0.524} & \textbf{0.479} & 0.013
                           & \textbf{0.309} & \textbf{0.627} & \textbf{0.603} & 0.059
                           \\
    \Xhline{0.5pt}
    
    \multirow{4}[0]{*}{3} 
         & FullSup-s       & 0.189 & 0.357 & 0.370 & 0.207
                           & 0.185 & 0.390 & 0.306 & 0.134
                           & 0.213 & 0.347 & 0.445 & 0.075
                           & 0.213 & 0.347 & 0.445 & 0.075
                           \\
          & TransFT        & 0.268 & 0.470 & 0.450 & 0.234
                           & 0.221 & 0.456 & 0.340 & 0.165
                           & 0.273 & 0.506 & 0.541 & 0.089
                           & 0.324 & 0.572 & 0.581 & 0.137
                           \\
          & SemiSup        & 0.208 & 0.371 & 0.396 & 0.216
                           & 0.196 & 0.413 & 0.316 & 0.145
                           & 0.230 & 0.373 & 0.476 & \textbf{0.092}
                           & 0.230 & 0.373 & 0.476 & 0.092
                           \\
          & SGFSIS (ours)   & \textbf{0.277} & \textbf{0.482} & \textbf{0.469} & \textbf{0.260}
                           & \textbf{0.316} & \textbf{0.574} & \textbf{0.428} & \textbf{0.239}
                           & \textbf{0.293} & \textbf{0.536} & \textbf{0.593} & 0.075
                           & \textbf{0.389} & \textbf{0.601} & \textbf{0.682} & \textbf{0.191}
                           \\
    \Xhline{0.5pt}
    
    \multirow{4}[0]{*}{5} 
          & FullSup-s      & 0.235 & 0.413 & 0.431 & 0.229
                           & 0.219 & 0.425 & 0.340 & 0.161
                           & 0.271 & 0.452 & 0.549 & 0.094
                           & 0.271 & 0.452 & 0.549 & 0.094
                           \\
          & TransFT        & 0.291 & 0.490 & 0.491 & 0.242
                           & 0.256 & 0.471 & 0.375 & 0.219
                           & 0.318 & 0.532 & 0.621 & 0.114
                           & 0.381 & 0.599 & 0.663 & 0.182
                           \\
          & SemiSup        & 0.252 & 0.430 & 0.454 & 0.227
                           & 0.240 & 0.447 & 0.361 & 0.189 
                           & 0.315 & 0.497 & 0.608 & 0.128
                           & 0.315 & 0.497 & 0.608 & 0.128
                           \\
          & SGFSIS (ours)   & \textbf{0.305} & \textbf{0.503} & \textbf{0.504} & \textbf{0.279}
                           & \textbf{0.356} & \textbf{0.585} & \textbf{0.465} & \textbf{0.298}
                           & \textbf{0.336} & \textbf{0.550} & \textbf{0.623} & \textbf{0.141}
                           & \textbf{0.422} & \textbf{0.608} & \textbf{0.726} & \textbf{0.228}
                           \\
    \Xhline{0.5pt}
    
    \multirow{4}[0]{*}{10} 
          & FullSup-s      & 0.263 & 0.439 & 0.488 & 0.271
                           & 0.273 & 0.476 & 0.386 & 0.214
                           & 0.314 & 0.482 & 0.630 & 0.151
                           & 0.314 & 0.482 & 0.630 & 0.151
                           \\
          & TransFT        & \textbf{0.314} & 0.501 & 0.508 & \textbf{0.306}
                           & 0.277 & 0.485 & 0.407 & 0.243
                           & 0.376 & 0.553 & 0.647 & 0.215
                           & 0.414 & 0.603 & 0.681 & 0.232
                           \\
          & SemiSup        & 0.291 & 0.461 & \textbf{0.527} & 0.294
                           & 0.294 & 0.502 & 0.419 & 0.241
                           & \textbf{0.393} & 0.548 & 0.670 & \textbf{0.227}
                           & 0.393 & 0.548 & 0.670 & 0.227
                           \\
          & SGFSIS (ours)   & 0.313 & \textbf{0.506} & 0.515 & 0.303
                           & \textbf{0.381} & \textbf{0.597} & \textbf{0.494} & \textbf{0.313}
                           & 0.376 & \textbf{0.555} & \textbf{0.672} & 0.193
                           & \textbf{0.453} & \textbf{0.617} & \textbf{0.733} & \textbf{0.295}
                           \\
    \Xhline{0.5pt}
    
    \multirow{4}[0]{*}{20} 
          & FullSup-s      & 0.310 & 0.469 & 0.516 & 0.321
                           & 0.340 & 0.509 & 0.463 & 0.294
                           & 0.395 & 0.528 & 0.679 & 0.235
                           & 0.395 & 0.528 & 0.679 & 0.235
                           \\
          & TransFT        & 0.338 & 0.510 & 0.546 & 0.337
                           & 0.310 & 0.550 & 0.435 & 0.294
                           & 0.410 & 0.558 & 0.693 & 0.245
                           & 0.445 & 0.601 & 0.695 & 0.294
                           \\
          & SemiSup        & 0.316 & 0.485 & 0.541 & 0.305
                           & 0.383 & 0.540 & 0.491 & 0.349
                           & 0.412 & \textbf{0.567} & \textbf{0.694} & 0.252
                           & 0.412 & 0.567 & 0.694 & 0.252
                           \\
          & SGFSIS (ours)   & \textbf{0.343} & \textbf{0.516} & \textbf{0.539} & \textbf{0.337}
                           & \textbf{0.434} & \textbf{0.605} & \textbf{0.531} & \textbf{0.399}
                           & \textbf{0.421} & \textbf{0.567} & 0.683 & \textbf{0.266}
                           & \textbf{0.480} & \textbf{0.609} & \textbf{0.744} & \textbf{0.345}
                           \\
    \Xhline{0.5pt}
    
    \multirow{4}[0]{*}{50} 
          & FullSup-s      & 0.340 & 0.504 & 0.552 & 0.332
                           & 0.423 & 0.544 & 0.524 & 0.419
                           & 0.466 & 0.591 & 0.720 & 0.341
                           & 0.466 & 0.591 & 0.720 & 0.341
                           \\
          & TransFT        & 0.345 & \textbf{0.518} & 0.540 & 0.357
                           & 0.354 & 0.518 & 0.470 & 0.351
                           & 0.445 & 0.590 & 0.714 & 0.312
                           & 0.487 & 0.619 & 0.713 & 0.372
                           \\
          & SemiSup        & 0.343 & 0.513 & \textbf{0.564} & 0.328
                           & 0.437 & 0.555 & 0.538 & 0.430
                           & \textbf{0.481} & \textbf{0.620} & \textbf{0.737} & \textbf{0.352}
                           & 0.481 & 0.620 & 0.737 & 0.352
                           \\
          & SGFSIS (ours)   & \textbf{0.353} & \textbf{0.518} & 0.559 & \textbf{0.362}
                           & \textbf{0.468} & \textbf{0.616} & \textbf{0.564} & \textbf{0.443}
                           & \textbf{0.481} & 0.603 & 0.700 & 0.334
                           & \textbf{0.504} & \textbf{0.626} & \textbf{0.757} & \textbf{0.401}
                           \\
    \Xhline{0.5pt}
    
    All   & FullSup & 0.382 & 0.531 & 0.591 & 0.381
                    & 0.453 & 0.585 & 0.553 & 0.469 
                    & 0.533 & 0.635 & 0.766 & 0.449
                    & 0.533 & 0.635 & 0.766 & 0.449
    \\
    \Xhline{1pt}

    \end{tabular}
    \label{tab:quanticomp}}
    \end{scriptsize}
\end{table*}

\subsection{Performance Metrics} 

\vspace{-2pt}

Three commonly-used performance metrics are adopted for quantitative evaluation and comparison from different perspectives, as follows

\begin{itemize}

\item \textbf{$\text{F}_\text{1}$-score} quantifies the performance of nucleus instance localization, as described in ~\cite{kumar2017dataset}. $\text{F}_\text{1}$-scores are firstly calculated over every single class independently and averaged over all the novel classes $\mathcal{C}$ and the overlapping classes  $\mathcal{C} \cap \mathcal{C}^{\text{base}}$ to give  \textbf{$\text{F}_\text{1}$-novel} and \textbf{$\text{F}_\text{1}$-base} respectively.
%, defined by
%\begin{equation}
%F_1 = \frac{2|TP|}{2|TP| + |FP| + |FN|},
%\end{equation}
%where $|TP|$, $|FP|$, $|FN|$ are the numbers of true positives, false positives and false negatives respectively, counted by matching the instance masks in the ground truth against the predictions and a match is admitted if their centers are closer than a threshold. 

\item \textbf{Aggregated Jaccard Index (AJI)} measures the quality of nucleus instance segmentation~\cite{kumar2017dataset}, i.e., the aggregated instance-wise concordance of spatial shapes between the ground truth and the prediction. AJI is calculated over true-positive instance pairs while being unaware of specific classes. We exactly follow~\cite{kumar2017dataset} for the definition and calculation of this metric.

\item \textbf{Multi-Class Panoptic Quality (mPQ)} combinatorially evaluates  the overall quality of nucleus localization and segmentation, which is defined in~\cite{kirillov2019panoptic,han2022meta}.  
%\begin{equation}
%{mPQ} = \frac{1}{N} \sum_{n=1}^N {PQ}_n,
%\end{equation}

%\begin{equation}
%PQ_n =  F_1 \times \underbrace{\frac{\sum\limits_{\left(p,
%q\right) \in TP_n} {IoU}\left(p, q\right)}{\left|TP_n\right|}}_{\text {Segmentation Quality}},
%\end{equation}
%where $PQ_n$ is the performance of the class $n$, $IoU$ the intersection-over-union operation and $TP_n$ the true-positive instance pairs of the class $n$. 
\end{itemize}
We use mPQ as the primary performance metric while also reporting AJI, $\text{F}_\text{1}$-novel and $\text{F}_\text{1}$-base to enable more in-depth analysis.

\subsection{Baselines for Comparison} \label{baseline sutup}

To our knowledge, there has been no previous work on applying FSL to nucleus instance segmentation that we can directly compare to. Hence, to enable comparative study, 
we construct three baselines as below. Notice that, for fair comparison, we use the same basic network structure as described in Section~\ref{sec:SGFSIS-framework}, with the guidance modules removed. Concretely, after the encoder and the decoder, a convolutional layer is added to each branch to generate the corresponding mask. 
\begin{itemize}

\item \textbf{FullSup-s}: Using only the support set $\mathcal{S}$ with labels to train the model in a fully-supervised way.

\item \textbf{TransFT}: Training a fully-supervised model on the labeled external dataset $\mathcal{D}^{\text{base}}$, and taking the support set with labels $\mathcal{S}$ to further fine-tune it before applying it to testing.

\item \textbf{SemiSup}: Adapting the well-known MeanTeacher algorithm~\cite{tarvainen2017mean} to train the model in a semi-supervised fashion, by the use of both the labeled samples $\mathcal{S}$ and the unlabeled samples $\mathcal{D}^u$. Note that consistency constraint is imposed on each of the four branches independently.
\end{itemize}
We also include the results obtained by using the entire target dataset $\mathcal{D}$ with labels to train the model in a fully-supervised way, termed as \textbf{FullSup}.

\subsection{Implementation Details} 
We implement the proposed SGFSIS framework with Pytorch (version 1.9) on a workstation with 2 NVIDIA RTX 3090 GPUs, each with 24 GB memory. The boundary label and the centroid label are converted from the instance label by using a disk filter (with the kernel size 3/5 and 0/3 for 20$\times$ and 40$\times$ magnification respectively). Data augmentation, including flip and rotation, are performed for our model and all the baseline methods. SGD optimizer is used, with the learning rate set to $10^{-4}$ for all the three training stages. Early stopping is imposed if there is no improvement for 10 consecutive epochs by screening the validation set. Considering the episode sampling is random, we repeat it for 5 times are reported the average performance for fair comparison.

\begin{figure}[!t]
	\centering
	\includegraphics[width = \linewidth]{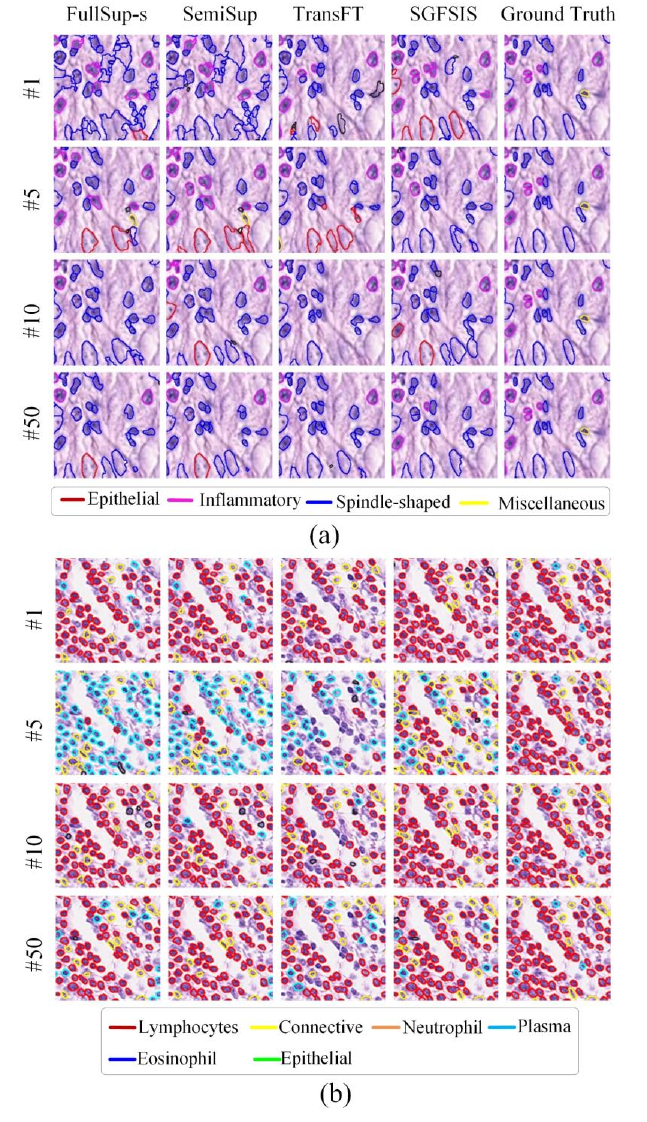}
	\caption{Qualitative comparison selected from the settings of (a) MoNuSAC $\rightarrow$ ConSep and (b) MoNuSAC $\rightarrow$ Lizard.}
\label{fig:quali-vis}
\end{figure}

\section{Results}
\label{sec:results}
%This section mainly presents experimental results and related analysis.

\subsection{Comparison with Baselines}
The quantitative results, in terms of four performance metrics, obtained by our SGFSIS and the three baselines under the four dataset settings are reported in Fig.~\ref{fig:quanti-results} and Table~\ref{tab:quanticomp}, where the number of shots varies by $K = \{1, 3, 5, 10, 20, 50\}$ for each setting. 

As can be observed, our SGFSIS can generally outperform the three compared baselines FullSup-s, TransFT and SemiSup, under the various settings of datasets and shot numbers (particularly smaller shot numbers), which validates its effectiveness. With the increase of the shot number $K$, the performance of SGFSIS can continuously improve and become very close or even better than the fully-supervised learning baseline FullSup when $K = 50$. Note that $K = 50$ is just less than 5$\%$ of the amount of FullSup's labels. This indicates, as a totally different paradigm, SGFSIS can indeed achieve annotation-efficient learning for nucleus instance segmentation like the other baselines.

SGFSIS outperforms more significantly in terms of AJI than $F_1$-novel/base, particularly when $K$ is very small.  This is because AJI reflects the quality of class-agnostic nucleus instance segmentation, which can hence benefit more from cross-data transfer than class-aware metrics like $F_1$-novel/base. Like our SGFSIS, TransFT also makes use of external datasets and significantly outperforms FullSup-s and SemiSup which rely only on the target dataset, but its performance is still worse than ours. This should be attributed to our carefully designed structural guidance mechanisms in SGFSIS. One can further observe that, the performance gain of our SGFSIS relative to FullSup-s and SemiSup is larger in terms of $F_1$-base than $F_1$-novel particularly in case of smaller $K$ values. This indicates the effectiveness of the generalization from FSL to GFSL.

It is also interesting to notice that the setting of datasets will affect the performance. For example, SGFSIS has larger performance gap relative to FullSup under the {ConSep $\rightarrow$ MoNuSAC} setting than under the {PanNuke $\rightarrow$ MoNuSAC} setting. This can be explained by that, CoNSep is a uni-center dataset while PanNuke is a multi-center, and an external dataset with larger diversity is more advantegous to transfer learning based methods including our SGFSIS.

Some representative qualitative results obtained by various methods are comparatively visualized in Fig.~\ref{fig:quali-vis}. As can be observed, SGFSIS is more advantageous in separating touching instances (see Fig.~\ref{fig:quali-vis} (a) and (b)). Also, it can perform better in novel-class classification.

\subsection{Ablation Study on the Structural Guidance Modules}

The superiority of our method in foreground instance segmentation stems from the introduction of the three structural guidance modules. To justify this point, we first conduct an ablation experiment, including two settings: 1) Removing one or two modules from the SGFSIS network while keeping the others unchanged; 2) Using a variant structure for the SGM modules, where we replace Eqn.~(\ref{eq:sgmMb}) with $
\boldsymbol{M}_B \gets \text{conv}_{\boldsymbol{\omega}_B}(
    \boldsymbol{F}_B^q) \otimes \boldsymbol{u}_B $.

\begin{figure}[!t]
	\centering
	\includegraphics[width = \linewidth]{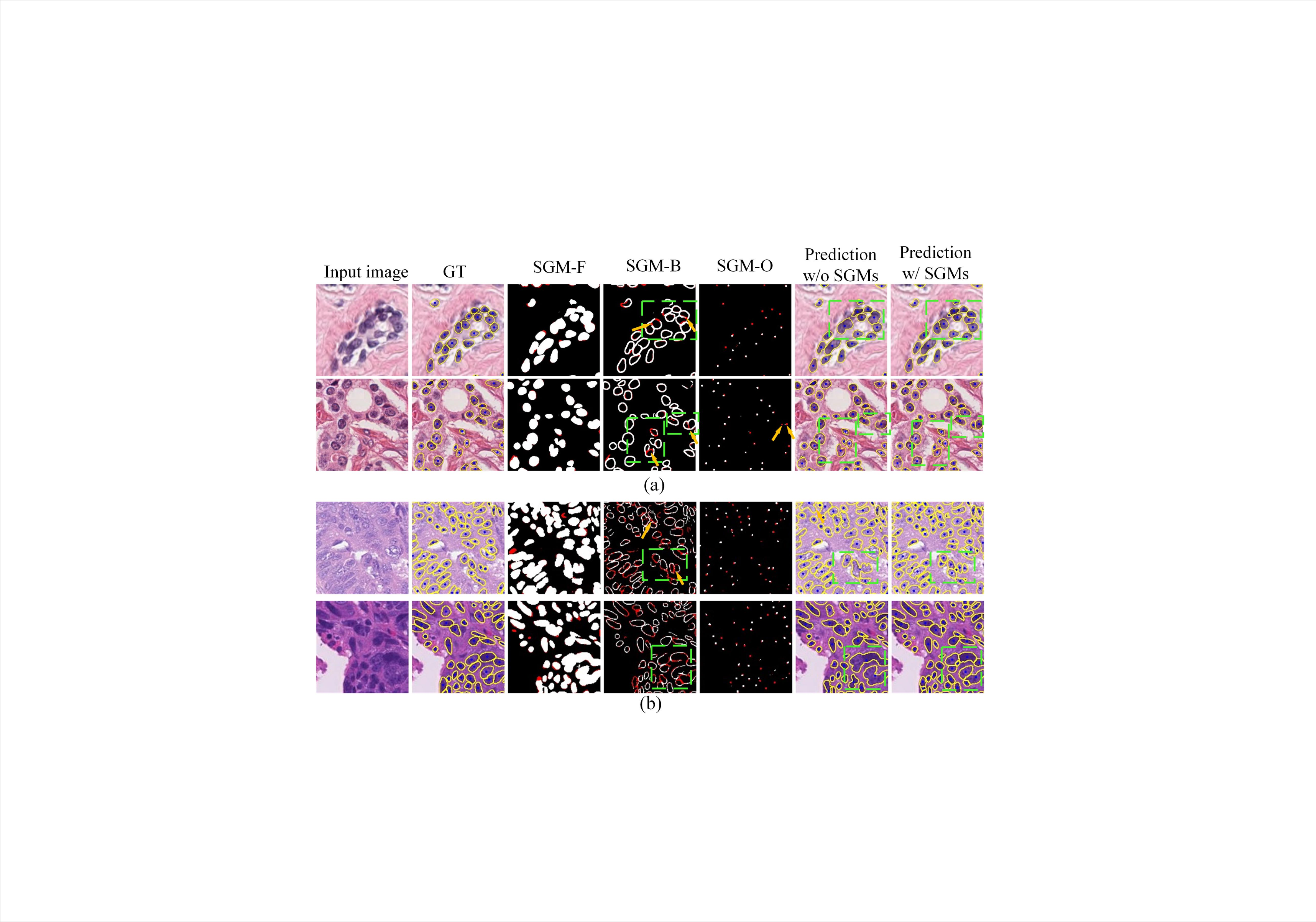}
	\caption{Representative results obtained by each SGM module as well as by using or without using the SGMs, under (a) MoNuSAC $\rightarrow$ PanNuke and (b) MoNuSAC $\rightarrow$ Lizard.} %For each picture, from left to right are original image with overlaid contour, predicted instance map, semantic segmenation map the predicted contour, from top to bottom are ground-truth, prediction without or with structural guidance. For prediction with structural guidance, the guided centroid and foreground are marked in red and blue respectively on the semantic segmenation map, while guided contour is marked in green on the contour prediction map.} 
\label{fig:guidance_compare} 
\end{figure}

The quantitative results under the settings of MoNuSAC $\rightarrow$ PanNuke and MoNuSAC $\rightarrow$ Lizard are reported in Table~\ref{tab:structguidance}. More intuitively, we illustrate in Fig.~\ref{fig:guidance_compare} some representative intermediate results obtained by our method, i.e., the results output by the three SGMs. As can be seen, all these modules contribute to the performance to some extent. 

\begin{table}[!t]
\scriptsize
\centering
\caption{Comparison with different configurations of the SGMs. * indicates using the variant SGM structure.}
\setlength{\tabcolsep}{1.6mm}{
\begin{tabular}{ccccccccc}
\Xhline{1pt}
\multicolumn{3}{c}{\textbf{SGMs}}
& \multicolumn{3}{c}{\textbf{MoNuSAC $\rightarrow$ PanNuke}} 
& \multicolumn{3}{c}{\textbf{MoNuSAC $\rightarrow$ Lizard}} \\ 
\cmidrule(lr){1-3}
\cmidrule(lr){4-6}
\cmidrule(lr){7-9}
SGM-F      & SGM-B     & SGM-O     & mPQ     & AJI     & Dice    & mPQ     & AJI     & Dice    
\\
\Xhline{1pt}
\multicolumn{9}{c}{1-shot}                                                    \\ \hline
                  &              &              & 0.211 & 0.474 & 0.649 & 0.243 & 0.521 & 0.748 \\
     $\checkmark$ &              &              & 0.226 & 0.498 & 0.682 & 0.247 & 0.528 & 0.755 \\
     $\checkmark$ & $\checkmark$ &              & 0.223 & 0.498 & 0.679 & 0.242 & 0.533 & 0.753 \\
     $\checkmark$ &              & $\checkmark$ & 0.221 & 0.501 & 0.681 & 0.250 & 0.535 & \textbf{0.756} \\
     $\checkmark$ & $\checkmark$ & $\checkmark$ & \textbf{0.230} & \textbf{0.525} & \textbf{0.701} & \textbf{0.256} & \textbf{0.546} & \textbf{0.756} \\
     $\checkmark^{*}$  & $\checkmark^{*}$ & $\checkmark^{*}$ & 0.206 & 0.488 & 0.671 & 0.227 & 0.509 & 0.742\\ \hline

\multicolumn{9}{c}{5-shot}                                                    \\ \hline
                  &              &              & 0.255 & 0.567 & 0.754 & 0.337 & 0.544 & 0.780\\
     $\checkmark$ &              &              & 0.281 & 0.582 & 0.767 & 0.348 & 0.550 & 0.784\\
     $\checkmark$ & $\checkmark$ &              & 0.281 & 0.585 & 0.769 & 0.350 & 0.582 & 0.785\\
     $\checkmark$ &              & $\checkmark$ & \textbf{0.285} & 0.586 & 0.766 & 0.351 & 0.563 & \textbf{0.786} \\
     $\checkmark$ & $\checkmark$ & $\checkmark$ & 0.279 & \textbf{0.591} & \textbf{0.769} & \textbf{0.356} & \textbf{0.585} & 0.782   \\
     $\checkmark^{*}$  & $\checkmark^{*}$ & $\checkmark^{*}$ & 0.274 & 0.579 & 0.760 & 0.320 & 0.547 &  0.776\\ \Xhline{1pt}
\end{tabular}}
\label{tab:structguidance}
\end{table}

\subsection{Validation of the Guided Classification Module}

The considerate design of the Guided Classification Module is very important to the effectiveness of our method. Here we further perform several ablation experiments for justification. To this end, we design the following two variants:

\begin{itemize}

\item \textbf{GCM-var\#1:}  Replacing the GCM module with a naive structure, i.e., simply using convolutional layers to predict foreground mask. 

\item \textbf{GCM-var\#2:} Removing the prototype registration procedure in the GCM module, i.e., without using Eq.~\ref{eq:protofuse} to fuse the prototypes with the base-class prototypes.     
\end{itemize}

The inferiority of GCM-var\#1 indicates the effectiveness of our designed GCM module. Compared to GCM-var\#2, our SGFSIS is better in terms of $F_1$-base, which reveals the prototype registration procedure can effectively prevent performance deterioration of the base classes.

\begin{table}[!t]
\scriptsize
  \centering
  \caption{Validation of the GCM module by comparing with two variants.}
  \setlength{\tabcolsep}{4.5pt}
\begin{tabular}{ ccccccc }
\Xhline{1pt}
\multicolumn{1}{c}{\multirow{2}{*}{\textbf{Methods}}} & \multicolumn{3}{c}{\textbf{MoNuSAC}$\rightarrow$\textbf{CoNSeP}} & \multicolumn{3}{c}{\textbf{MoNuSAC}$\rightarrow$\textbf{Lizard}} \\ 
\cmidrule(lr){2-4}
\cmidrule(lr){5-7}
\multicolumn{1}{c}{}                        & mPQ  & $F_1$-base  & $F_1$-novel  & mPQ  & $F_1$-base  &$F_1$-novel  \\ \Xhline{1pt}
\multicolumn{7}{c}{1-shot}                                                                       \\ \hline
GCM-var\#1                                  & 0.229 & 0.413 & 0.165 & 0.241 & 0.341 & 0.151     \\
GCM-var\#2                                  & 0.241 & 0.441 & \textbf{0.170} & 0.253 & 0.360 & 0.157     \\
SGFSIS (ours)                                & \textbf{0.244} & \textbf{0.452} & 0.167 & \textbf{0.256} & \textbf{0.367} & \textbf{0.158}     \\ \hline
\multicolumn{7}{c}{5-shot}                                                                       \\ \hline
GCM-var\#1                                  & 0.287 & 0.486 & 0.255 & 0.336 & 0.431 & 0.269     \\
GCM-var\#2                                  & 0.295 & 0.493 & \textbf{0.280} & 0.350 & 0.439 & 0.294     \\
SGFSIS (ours)                                & \textbf{0.305} & \textbf{0.504} & 0.279 & \textbf{0.356} & \textbf{0.465} & \textbf{0.298}     \\ \Xhline{1pt}
\end{tabular}
\label{tab:guidedclassificationmodule}
\end{table}

\section{Discussion and Conclusions}
\label{sec:concl}
The extremely laborious and expertise-dependent annotation hampers the construction of high-accuracy nucleus instance segmentation models, which is becoming the bottleneck of many computational pathology applications. To break through this bottleneck, previous works have investigated a couple of annotation-efficient learning paradigms, like semi-supervised learning, generative adversarial learning, domain adaptation, etc. Despite the advances to some extent, there still exists pressing demand on the study of more effective annotation-efficient learning paradigms for nucleus instance segmentation. 

Our primary contribution is to introduce FSL as an annotation-efficient learning paradigm, motivated by the fact an increasing number of fully-annotated public datasets are emerging recently. It is stressed that, FSL was originally developed to tackle the scarcity of data, rather than annotation. We innovatively adapt FSL to the task of nucleus instance segmentation where the core challenge is the scarcity of annotation, which is the first attempt to our knowledge. Such adaptation is by no means trivial, with a couple of issues to be addressed, like coping with the overlapping classes between the target dataset and the external dataset, designing the structural guidance mechanism, etc. We have provided an effective framework to achieve this goal. 

As for the experimental results, we highlight two points: 1) Our SGFSIS and the three baseline methods can achieve comparable performance (mostly with a gap less than 5\%)  to FullSup, with less than 5\% of FullSup's annotation. These results are inline with previous works~\cite{wu2022cross}. It implies that, for the task of nucleus instance segmentation,  full annotation of numerous training samples might be redundant and unnecessary and how to make more efficient use of the annotation  has large room for exploration. 2) The much larger superiority of our SGFSIS in terms of AJI than $F_1$-novel/base indicates that cross-dataset transfer is easier for the sub-task of nucleus boundary localization than the sub-task of nucleus classification. This suggests that these two sub-tasks might be better decoupled and resolved separately.    

Our proposed method inevitably has several limitations. First, our method by definition depends on the accessibility to a fully-annotated external dataset. It hence cannot be applied if such a condition is not satisfied. Second, by our observation, nuclei with very low visual contrast to the background, which may be caused by the low quality of slide staining or digital scanning, will be missed by our method. Third, the current work only makes use of the limited labeled data of the target dataset and the fully-annotated external dataset, while totally ignoring the unlabeled data of the target dataset which is actually very valuable to the improvement of an annotation-efficient learning method. In our future work, we will further explore along these directions.  

\section*{Acknowledgments}
This work was funded by the National Key R\&D Program of China under Grant No. 2021YFF1201004, the Natural Science Foundation of China under Grants No. 62076099, 82273358 and 82003059, and the Guangzhou Basic and Applied Basic Research Topics under Grant No. 2023A04J2383.

\bibliographystyle{IEEEtran}
\bibliography{IEEEabrv,ref}

\begin{thebibliography}{10}
\providecommand{\url}[1]{#1}
\csname url@samestyle\endcsname
\providecommand{\newblock}{\relax}
\providecommand{\bibinfo}[2]{#2}
\providecommand{\BIBentrySTDinterwordspacing}{\spaceskip=0pt\relax}
\providecommand{\BIBentryALTinterwordstretchfactor}{4}
\providecommand{\BIBentryALTinterwordspacing}{\spaceskip=\fontdimen2\font plus
\BIBentryALTinterwordstretchfactor\fontdimen3\font minus \fontdimen4\font\relax}
\providecommand{\BIBforeignlanguage}[2]{{%
\expandafter\ifx\csname l@#1\endcsname\relax
\typeout{** WARNING: IEEEtran.bst: No hyphenation pattern has been}%
\typeout{** loaded for the language `#1'. Using the pattern for}%
\typeout{** the default language instead.}%
\else
\language=\csname l@#1\endcsname
\fi
#2}}
\providecommand{\BIBdecl}{\relax}
\BIBdecl

\bibitem{abels2019computational}
E.~Abels, L.~Pantanowitz, F.~Aeffner, M.~D. Zarella, J.~Van Der~Laak, M.~M. Bui, V.~N. Vemuri, A.~V. Parwani, J.~Gibbs, E.~Agosto-Arroyo \emph{et~al.}, ``Computational pathology definitions, best practices, and recommendations for regulatory guidance: a white paper from the digital pathology association,'' \emph{The Journal of Pathology}, vol. 249, no.~3, pp. 286--294, 2019.

\bibitem{yuan2012quantitative}
Y.~Yuan, H.~Failmezger, O.~M. Rueda, H.~R. Ali, S.~Gr{\"a}f, S.-F. Chin, R.~F. Schwarz, C.~Curtis, M.~J. Dunning, H.~Bardwell \emph{et~al.}, ``Quantitative image analysis of cellular heterogeneity in breast tumors complements genomic profiling,'' \emph{Science Translational Medicine}, vol.~4, no. 157, pp. 143--153, 2012.

\bibitem{kapil2021domain}
A.~Kapil, A.~Meier, K.~Steele, M.~Rebelatto, K.~Nekolla, A.~Haragan, A.~Silva, A.~Zuraw, C.~Barker, M.~L. Scott \emph{et~al.}, ``Domain adaptation-based deep learning for automated {Tumor Cell (TC)} scoring and survival analysis on pd-l1 stained tissue images,'' \emph{IEEE Transactions on Medical Imaging}, vol.~40, no.~9, pp. 2513--2523, 2021.

\bibitem{qaiser2019learning}
T.~Qaiser and N.~M. Rajpoot, ``Learning where to see: A novel attention model for automated immunohistochemical scoring,'' \emph{IEEE Transactions on Medical Imaging}, vol.~38, no.~11, pp. 2620--2631, 2019.

\bibitem{gurcan2009histopathological}
M.~N. Gurcan, L.~E. Boucheron, A.~Can, A.~Madabhushi, N.~M. Rajpoot, and B.~Yener, ``Histopathological image analysis: A review,'' \emph{IEEE Reviews in Biomedical Engineering}, vol.~2, pp. 147--171, 2009.

\bibitem{li2020self}
Y.~Li, J.~Chen, X.~Xie, K.~Ma, and Y.~Zheng, ``Self-loop uncertainty: A novel pseudo-label for semi-supervised medical image segmentation,'' in \emph{Proceedings of the International Conference on Medical Image Computing and Computer-Assisted Intervention}, 2020, pp. 614--623.

\bibitem{wu2022cross}
H.~Wu, Z.~Wang, Y.~Song, L.~Yang, and J.~Qin, ``Cross-patch dense contrastive learning for semi-supervised segmentation of cellular nuclei in histopathologic images,'' in \emph{Proceedings of the IEEE Conference on Computer Vision and Pattern Recognition}, 2022, pp. 11\,666--11\,675.

\bibitem{jin2022semi}
Q.~Jin, H.~Cui, C.~Sun, J.~Zheng, L.~Wei, Z.~Fang, Z.~Meng, and R.~Su, ``Semi-supervised histological image segmentation via hierarchical consistency enforcement,'' in \emph{Proceedings of the International Conference on Medical Image Computing and Computer-Assisted Intervention}, 2022, pp. 3--13.

\bibitem{hou2019robust}
L.~Hou, A.~Agarwal, D.~Samaras, T.~M. Kurc, R.~R. Gupta, and J.~H. Saltz, ``Robust histopathology image analysis: To label or to synthesize?'' in \emph{Proceedings of the IEEE Conference on Computer Vision and Pattern Recognition}, 2019, pp. 8533--8542.

\bibitem{mahmood2019deep}
F.~Mahmood, D.~Borders, R.~J. Chen, G.~N. McKay, K.~J. Salimian, A.~Baras, and N.~J. Durr, ``Deep adversarial training for multi-organ nuclei segmentation in histopathology images,'' \emph{IEEE Transactions on Medical Imaging}, vol.~39, no.~11, pp. 3257--3267, 2019.

\bibitem{graham2019hover}
S.~Graham, Q.~D. Vu, S.~E.~A. Raza, A.~Azam, Y.~W. Tsang, J.~T. Kwak, and N.~Rajpoot, ``Hover-net: Simultaneous segmentation and classification of nuclei in multi-tissue histology images,'' \emph{Medical Image Analysis}, vol.~58, no. 101563, 2019.

\bibitem{gamper2019pannuke}
J.~Gamper, N.~Alemi~Koohbanani, K.~Benet, A.~Khuram, and N.~Rajpoot, ``Pannuke: an open pan-cancer histology dataset for nuclei instance segmentation and classification,'' in \emph{Proceedings of the European Congress on Digital Pathology}, 2019, pp. 11--19.

\bibitem{verma2021monusac2020}
R.~Verma, N.~Kumar, A.~Patil, N.~C. Kurian, S.~Rane, S.~Graham, Q.~D. Vu, M.~Zwager, S.~E.~A. Raza, N.~Rajpoot \emph{et~al.}, ``{MoNuSAC2020}: A multi-organ nuclei segmentation and classification challenge,'' \emph{IEEE Transactions on Medical Imaging}, vol.~40, no.~12, pp. 3413--3423, 2021.

\bibitem{liu2020unsupervised}
D.~Liu, D.~Zhang, Y.~Song, F.~Zhang, L.~O'Donnell, H.~Huang, M.~Chen, and W.~Cai, ``Unsupervised instance segmentation in microscopy images via panoptic domain adaptation and task re-weighting,'' in \emph{Proceedings of the IEEE Conference on Computer Vision and Pattern Recognition}, 2020, pp. 4243--4252.

\bibitem{li2022domain}
C.~Li, D.~Liu, H.~Li, Z.~Zhang, G.~Lu, X.~Chang, and W.~Cai, ``Domain adaptive nuclei instance segmentation and classification via category-aware feature alignment and pseudo-labelling,'' in \emph{Proceedings of the International Conference on Medical Image Computing and Computer-Assisted Intervention}, 2022, pp. 715--724.

\bibitem{yang2021minimizing}
S.~Yang, J.~Zhang, J.~Huang, B.~C. Lovell, and X.~Han, ``Minimizing labeling cost for nuclei instance segmentation and classification with cross-domain images and weak labels,'' in \emph{Proceedings of the Association for the Advancement of Artificial Intelligence}, vol.~35, no.~1, 2021, pp. 697--705.

\bibitem{wang2020generalizing}
Y.~Wang, Q.~Yao, J.~T. Kwok, and L.~M. Ni, ``Generalizing from a few examples: A survey on few-shot learning,'' \emph{ACM Computing Surveys}, vol.~53, no.~3, pp. 1--34, 2020.

\bibitem{vinyals2016matching}
O.~Vinyals, C.~Blundell, T.~Lillicrap, D.~Wierstra \emph{et~al.}, ``Matching networks for one shot learning,'' in \emph{Proceedings of the Advances in Neural Information Processing Systems}, 2016, pp. 3630--3638.

\bibitem{lang2023base}
C.~Lang, G.~Cheng, B.~Tu, C.~Li, and J.~Han, ``Base and meta: A new perspective on few-shot segmentation,'' \emph{IEEE Transactions on Pattern Analysis and Machine Intelligence}, 2023.

\bibitem{graham2021lizard}
S.~Graham, M.~Jahanifar, A.~Azam, M.~Nimir, Y.-W. Tsang, K.~Dodd, E.~Hero, H.~Sahota, A.~Tank, K.~Benes \emph{et~al.}, ``Lizard: a large-scale dataset for colonic nuclear instance segmentation and classification,'' in \emph{Proceedings of the IEEE International Conference on Computer Vision}, 2021, pp. 684--693.

\bibitem{ronneberger2015u}
O.~Ronneberger, P.~Fischer, and T.~Brox, ``U-{N}et: Convolutional networks for biomedical image segmentation,'' in \emph{Proceedings of the International Conference on Medical Image Computing and Computer-Assisted Intervention}, 2015, pp. 234--241.

\bibitem{raza2019micro}
S.~E.~A. Raza, L.~Cheung, M.~Shaban, S.~Graham, D.~Epstein, S.~Pelengaris, M.~Khan, and N.~M. Rajpoot, ``Micro-{N}et: A unified model for segmentation of various objects in microscopy images,'' \emph{Medical Image Analysis}, vol.~52, pp. 160--173, 2019.

\bibitem{qu2019improving}
H.~Qu, Z.~Yan, G.~M. Riedlinger, S.~De, and D.~N. Metaxas, ``Improving nuclei/gland instance segmentation in histopathology images by full resolution neural network and spatial constrained loss,'' in \emph{Proceedings of the International Conference on Medical Image Computing and Computer-Assisted Intervention}, 2019, pp. 378--386.

\bibitem{xing2015automatic}
F.~Xing, Y.~Xie, and L.~Yang, ``An automatic learning-based framework for robust nucleus segmentation,'' \emph{IEEE Transactions on Medical Imaging}, vol.~35, no.~2, pp. 550--566, 2015.

\bibitem{chen2017dcan}
H.~Chen, X.~Qi, L.~Yu, Q.~Dou, J.~Qin, and P.-A. Heng, ``{DCAN}: Deep contour-aware networks for object instance segmentation from histology images,'' \emph{Medical Image Analysis}, vol.~36, pp. 135--146, 2017.

\bibitem{ke2023clusterseg}
J.~Ke, Y.~Lu, Y.~Shen, J.~Zhu, Y.~Zhou, J.~Huang, J.~Yao, X.~Liang, Y.~Guo, Z.~Wei \emph{et~al.}, ``Clusterseg: A crowd cluster pinpointed nucleus segmentation framework with cross-modality datasets,'' \emph{Medical Image Analysis}, vol.~85, no. 102758, 2023.

\bibitem{zhou2019cia}
Y.~Zhou, O.~F. Onder, Q.~Dou, E.~Tsougenis, H.~Chen, and P.-A. Heng, ``{CIA-Net}: Robust nuclei instance segmentation with contour-aware information aggregation,'' in \emph{Proceedings of the International Conference on Information Processing in Medical Imaging}, 2019, pp. 682--693.

\bibitem{zhao2020triple}
B.~Zhao, X.~Chen, Z.~Li, Z.~Yu, S.~Yao, L.~Yan, Y.~Wang, Z.~Liu, C.~Liang, and C.~Han, ``Triple {U-Net}: Hematoxylin-aware nuclei segmentation with progressive dense feature aggregation,'' \emph{Medical Image Analysis}, vol.~65, no. 101786, 2020.

\bibitem{naylor2018segmentation}
P.~Naylor, M.~La{\'e}, F.~Reyal, and T.~Walter, ``Segmentation of nuclei in histopathology images by deep regression of the distance map,'' \emph{IEEE Transactions on Medical Imaging}, vol.~38, no.~2, pp. 448--459, 2018.

\bibitem{he2021cdnet}
H.~He, Z.~Huang, Y.~Ding, G.~Song, L.~Wang, Q.~Ren, P.~Wei, Z.~Gao, and J.~Chen, ``{CDNet}: Centripetal direction network for nuclear instance segmentation,'' in \emph{Proceedings of the IEEE International Conference on Computer Vision}, 2021, pp. 4026--4035.

\bibitem{chen2023cpp}
S.~Chen, C.~Ding, M.~Liu, J.~Cheng, and D.~Tao, ``{CPP-Net}: Context-aware polygon proposal network for nucleus segmentation,'' \emph{IEEE Transactions on Image Processing}, vol.~32, pp. 980--994, 2023.

\bibitem{he2023toposeg}
H.~He, J.~Wang, P.~Wei, F.~Xu, X.~Ji, C.~Liu, and J.~Chen, ``{TopoSeg}: Topology-aware nuclear instance segmentation,'' in \emph{Proceedings of the IEEE International Conference on Computer Vision}, 2023, pp. 21\,307--21\,316.

\bibitem{qu2020weakly}
H.~Qu, P.~Wu, Q.~Huang, J.~Yi, Z.~Yan, K.~Li, G.~M. Riedlinger, S.~De, S.~Zhang, and D.~N. Metaxas, ``Weakly supervised deep nuclei segmentation using partial points annotation in histopathology images,'' \emph{IEEE Transactions on Medical Imaging}, vol.~39, no.~11, pp. 3655--3666, 2020.

\bibitem{lin2024bonus}
Y.~Lin, Z.~Wang, D.~Zhang, K.-T. Cheng, and H.~Chen, ``{BoNuS}: Boundary mining for nuclei segmentation with partial point labels,'' \emph{IEEE Transactions on Medical Imaging}, 2024.

\bibitem{zhou2023cyclic}
Y.~Zhou, Y.~Wu, Z.~Wang, B.~Wei, M.~Lai, J.~Shou, Y.~Fan, and Y.~Xu, ``Cyclic learning: Bridging image-level labels and nuclei instance segmentation,'' \emph{IEEE Transactions on Medical Imaging}, 2023.

\bibitem{lou2022pixel}
W.~Lou, H.~Li, G.~Li, X.~Han, and X.~Wan, ``Which pixel to annotate: A label-efficient nuclei segmentation framework,'' \emph{IEEE Transactions on Medical Imaging}, 2022.

\bibitem{han2022meta}
C.~Han, H.~Yao, B.~Zhao, Z.~Li, Z.~Shi, L.~Wu, X.~Chen, J.~Qu, K.~Zhao, R.~Lan \emph{et~al.}, ``Meta multi-task nuclei segmentation with fewer training samples,'' \emph{Medical Image Analysis}, vol.~80, no. 102481, 2022.

\bibitem{fan2020fgn}
Z.~Fan, J.-G. Yu, Z.~Liang, J.~Ou, C.~Gao, G.-S. Xia, and Y.~Li, ``{FGN}: Fully guided network for few-shot instance segmentation,'' in \emph{Proceedings of the IEEE Conference on Computer Vision and Pattern Recognition}, 2020, pp. 9172--9181.

\bibitem{wang2022dynamic}
H.~Wang, J.~Liu, Y.~Liu, S.~Maji, J.-J. Sonke, and E.~Gavves, ``Dynamic transformer for few-shot instance segmentation,'' in \emph{Proceedings of the ACM International Conference on Multimedia}, 2022, pp. 2969--2977.

\bibitem{kohler2023few}
M.~K{\"o}hler, M.~Eisenbach, and H.-M. Gross, ``Few-shot object detection: A comprehensive survey,'' \emph{IEEE Transactions on Neural Networks and Learning Systems}, 2023.

\bibitem{fan2021generalized}
Z.~Fan, Y.~Ma, Z.~Li, and J.~Sun, ``Generalized few-shot object detection without forgetting,'' in \emph{Proceedings of the IEEE Conference on Computer Vision and Pattern Recognition}, 2021, pp. 4527--4536.

\bibitem{ma2023digeo}
J.~Ma, Y.~Niu, J.~Xu, S.~Huang, G.~Han, and S.-F. Chang, ``{DiGeo}: Discriminative geometry-aware learning for generalized few-shot object detection,'' in \emph{Proceedings of the IEEE Conference on Computer Vision and Pattern Recognition}, 2023, pp. 3208--3218.

\bibitem{roy2020squeeze}
A.~G. Roy, S.~Siddiqui, S.~P{\"o}lsterl, N.~Navab, and C.~Wachinger, ``‘{Squeeze} \& excite’guided few-shot segmentation of volumetric images,'' \emph{Medical Image Analysis}, vol.~59, no. 101587, 2020.

\bibitem{cui2020unified}
H.~Cui, D.~Wei, K.~Ma, S.~Gu, and Y.~Zheng, ``A unified framework for generalized low-shot medical image segmentation with scarce data,'' \emph{IEEE Transactions on Medical Imaging}, vol.~40, no.~10, pp. 2656--2671, 2020.

\bibitem{tang2021recurrent}
H.~Tang, X.~Liu, S.~Sun, X.~Yan, and X.~Xie, ``Recurrent mask refinement for few-shot medical image segmentation,'' in \emph{Proceedings of the IEEE International Conference on Computer Vision}, 2021, pp. 3918--3928.

\bibitem{feng2023learning}
Y.~Feng, Y.~Wang, H.~Li, M.~Qu, and J.~Yang, ``Learning what and where to segment: A new perspective on medical image few-shot segmentation,'' \emph{Medical Image Analysis}, vol.~87, no. 102834, 2023.

\bibitem{tian2022generalized}
Z.~Tian, X.~Lai, L.~Jiang, S.~Liu, M.~Shu, H.~Zhao, and J.~Jia, ``Generalized few-shot semantic segmentation,'' in \emph{Proceedings of the IEEE Conference on Computer Vision and Pattern Recognition}, 2022, pp. 11\,563--11\,572.

\bibitem{liu2023learning}
S.-A. Liu, Y.~Zhang, Z.~Qiu, H.~Xie, Y.~Zhang, and T.~Yao, ``Learning orthogonal prototypes for generalized few-shot semantic segmentation,'' in \emph{Proceedings of the IEEE Conference on Computer Vision and Pattern Recognition}, 2023, pp. 11\,319--11\,328.

\bibitem{hospedales2021meta}
T.~Hospedales, A.~Antoniou, P.~Micaelli, and A.~Storkey, ``Meta-learning in neural networks: A survey,'' \emph{IEEE Transactions on Pattern Analysis and Machine Intelligence}, vol.~44, no.~9, pp. 5149--5169, 2021.

\bibitem{he2016deep}
K.~He, X.~Zhang, S.~Ren, and J.~Sun, ``Deep residual learning for image recognition,'' in \emph{Proceedings of the IEEE Conference on Computer Vision and Pattern Recognition}, 2016, pp. 770--778.

\bibitem{gamper2020pannuke}
J.~Gamper, N.~A. Koohbanani, K.~Benes, S.~Graham, M.~Jahanifar, S.~A. Khurram, A.~Azam, K.~Hewitt, and N.~Rajpoot, ``Pannuke dataset extension, insights and baselines,'' \emph{arXiv:2003.10778}, 2020.

\bibitem{kumar2017dataset}
N.~Kumar, R.~Verma, S.~Sharma, S.~Bhargava, A.~Vahadane, and A.~Sethi, ``A dataset and a technique for generalized nuclear segmentation for computational pathology,'' \emph{IEEE Transactions on Medical Imaging}, vol.~36, no.~7, pp. 1550--1560, 2017.

\bibitem{kirillov2019panoptic}
A.~Kirillov, K.~He, R.~Girshick, C.~Rother, and P.~Doll{\'a}r, ``Panoptic segmentation,'' in \emph{Proceedings of the IEEE Conference on Computer Vision and Pattern Recognition}, 2019, pp. 9404--9413.

\bibitem{tarvainen2017mean}
A.~Tarvainen and H.~Valpola, ``Mean teachers are better role models: Weight-averaged consistency targets improve semi-supervised deep learning results,'' \emph{Advances in Neural Information Processing Systems}, vol.~30, 2017.

\end{thebibliography}

\end{document}